\documentclass{article}

    \usepackage[numbers,sort]{natbib}

\usepackage[preprint]{neurips_2024}

\usepackage[utf8]{inputenc} 
\usepackage[T1]{fontenc}    
\usepackage{hyperref}       
\usepackage{url}            
\usepackage{booktabs}       
\usepackage{amsfonts}       
\usepackage{nicefrac}       
\usepackage{microtype}      

\usepackage{amsmath}
\DeclareMathOperator*{\argmax}{arg\,max}

\usepackage{bbm}
\usepackage{graphicx}

\usepackage{multirow}
\usepackage{booktabs}
\usepackage{wrapfig}
\usepackage{subfig}

\usepackage{caption} 
\usepackage[table,xcdraw]{xcolor}

\usepackage{mathtools}

\usepackage{bbm}

\title{Let the Fuzzy Rule Speak: Enhancing In-context Learning Debiasing with Interpretability}

\author{ Ruixi Lin \qquad Yang You\\
  Department of Computer Science\\
  National University of Singapore\\
  \texttt{\{ruixi,youy\}@comp.nus.edu.sg} \\}

\begin{document}

\maketitle

\begin{abstract}
Large language models (LLMs) often struggle with balanced class accuracy in text classification tasks using in-context learning (ICL), hindering some practical uses due to user dissatisfaction or safety risks caused by misclassifications. Retraining LLMs to address root causes in data or model priors is neither easy nor cost-effective. This paper delves deeper into the class accuracy imbalance issue, identifying that it arises because certain classes consistently receive disproportionately high ICL probabilities, causing under-prediction and lower accuracy for others. More importantly, probability ranges affect the imbalance differently, allowing for precise, range-specific corrections. We introduce FuRud (\textbf{Fu}zzy \textbf{Ru}le Optimization-based \textbf{D}ebiasing), a method for sample-level class probability correction. FuRud tackles interpretability challenges by determining why certain classes need corrections and tailoring adjustments for each instance's class probabilities which is powered by fuzzy sets with triangular membership functions, transforming a class probability based on the range it belongs to. By solving a nonlinear integer programming problem with a labeled set of ICL class probabilities to minimize class accuracy bias (COBias) and maximize overall accuracy, each class selects an optimal correction function from 19 triangular membership functions without updating an LLM, and the selected functions correct test instances at inference. Across seven benchmark datasets, FuRud reduces COBias by over half (56\%) and improves overall accuracy by 21\% relatively, outperforming state-of-the-art debiasing methods.
\end{abstract}

\section{Introduction}
\label{sec:intro}
In-context learning (ICL) allows large language models (LLMs) to perform text classification tasks by prompting them with a few demonstrative examples. However, the class accuracies are often imbalanced due to biases in the training data or model priors. Addressing such imbalance while improving overall accuracy is a compelling frontier in the realm of \textit{debiasing}. The skewness in the output space can be alleviated by inference-time corrections on ICL output logits or probabilities. For example, \citet{cobias} explicitly targets mitigating class accuracy differences and quantify them as COBias, the averaged pairwise class accuracy differences, and learn class-level correction weights. While effective, the prior method corrects any instance with the same set of correction weights, lacking considerations in capturing per-sample, per-class nuances.

A direct cause of the imbalance is that ICL often yields specific ranges of probabilities to each output class. Some classes receive high probabilities for any input, while others may not. The consequence is that the latter is less frequently chosen than the former, resulting in lower accuracies. On the sample level, for all instances of a ground-truth class \textit{A}, it is a general observation that instances with a low output probability in class \textit{A} will have lower accuracy compared to those instances with a higher output probability in class \textit{A}. The latter instances may not need as much amplification in class \textit{A} output probability as the former instances. This suggests that sample-level customized correction should be enabled to accommodate different ICL probability ranges within a same output class.

Therefore, we aim for a sample-level correction method that interpretively amplifies or reduces different ranges of an output class's probabilities. In this work, we address the pressing need for enhanced understandings in how biased ICL predictions happen with the following research questions, and propose a per-sample, per-class correction method using fuzzy representation techniques.

\noindent \textbf{RQ1: What is the interpretability challenge in correcting in-context learned representations?}
Given an $N$-class classification dataset, we denote the $m$-th instance's input prompt and class as $(x_m, y_m)$, where $x_m$ consists of a task instruction, few-shot demonstrative examples, the input text and a question on its class. An LLM in-context learns output class probabilities $\boldsymbol{p}_m=(p_{m1},\dots, p_{mN})$ (normalized over $N$ classes), then the prediction $\hat{y}_m$ is $\argmax_i p_{mi}$. The $\boldsymbol{p}_m$ may need corrections in one or more of the classes, to reduce imbalance in class accuracy and improve overall accuracy. The interpretability challenges raised in this process can be specified as (1) detecting which classes need corrections, and (2) for each correction-needed class, applying range-specific amplifications/reductions.

\noindent \textbf{RQ2: How can we achieve interpretable corrections with fuzzy rules?}
In short, we leverage membership functions to achieve interpretable corrections. For more backgrounds, interpretable machine learning systems need a human-readable subset of input attributes to generate the target \citep{Jethani2021,Carvalho2019}, so they often use fuzzy rules and fuzzy memberships, which provide interpretable quantifiers of given attributes (such as the size, \textit{Small}), to learn these systems \citep{Vernon2024,Vilone2020,fuzzy2007}. In classical fuzzy rule classification systems, input attributes are assigned to fuzzy sets to generate rules for pattern classification \citep{fuzzy1999,fuzzy2005,fuzzy2016,rudzinski2016multi,fuzzy2017Marian}. A fuzzy classification system contains multiple human-readable rules, which can be as simple as ``1. If attribute Bare Nuclei is \textit{Small} then the consequent (predicted) class is \textit{Benign}.2.\textit{...}3. If attribute Uniformity of Cell Size is \textit{not Small} then the consequent class is \textit{Malignant}.'' \citep{fuzzy2017Marian}. Here, \textit{Small} and \textit{not Small} are fuzzy sets, and their corresponding membership functions quantify the level of participation of each given attribute in the respective fuzzy set.

In this work, we leverage the range-wise transformation capabilities of membership functions for debiasing. A membership function is a curve that maps an input attribute to a fuzzy value between 0 and 1 \citep{fuzzysets}. Viewing class probabilities as input attributes, we can use membership functions to adjust the probabilities, as long as the membership functions are selected under debiasing objectives. The key intuition is that a membership function can asymmetrically amplify or reduce different ranges of inputs. As such, a fuzzy rule based debiaser is applied to $p_{mi}$, denoted as $f_{A_i}(p_{mi})$, where $A_i$ is a fuzzy set for class $i$, and its membership function $f_{A_i}$ maps $p_{mi}$ to a corrected $p'_{mi} := f_{A_i}(p_{mi})$. The debiaser can be viewed as a \textbf{single rule}:
\begin{equation}
        \text{If} \underbrace{\text{ class $1$ is $A_{1}$ and ... and class $N$ is $A_{N}$}}_\text{Antecedent}
        \text{ then }\underbrace{\text{predict $\textstyle \argmax_j f_{A_j}(p_{mj})$}}_\text{Consequent}
        \label{eq:1}
\end{equation}
Our goal is to optimize the selection of membership functions towards mitigating COBias and improving overall accuracy. Specially, we include a \textit{Don't Change} membership function that will keep a class unchanged. When a correction is needed, a piece of the triangular function is activated for evaluating the corrected probability based on the range that the input probability belongs to.

\begin{figure*}[ht]
    \centering
\includegraphics[width=\linewidth]{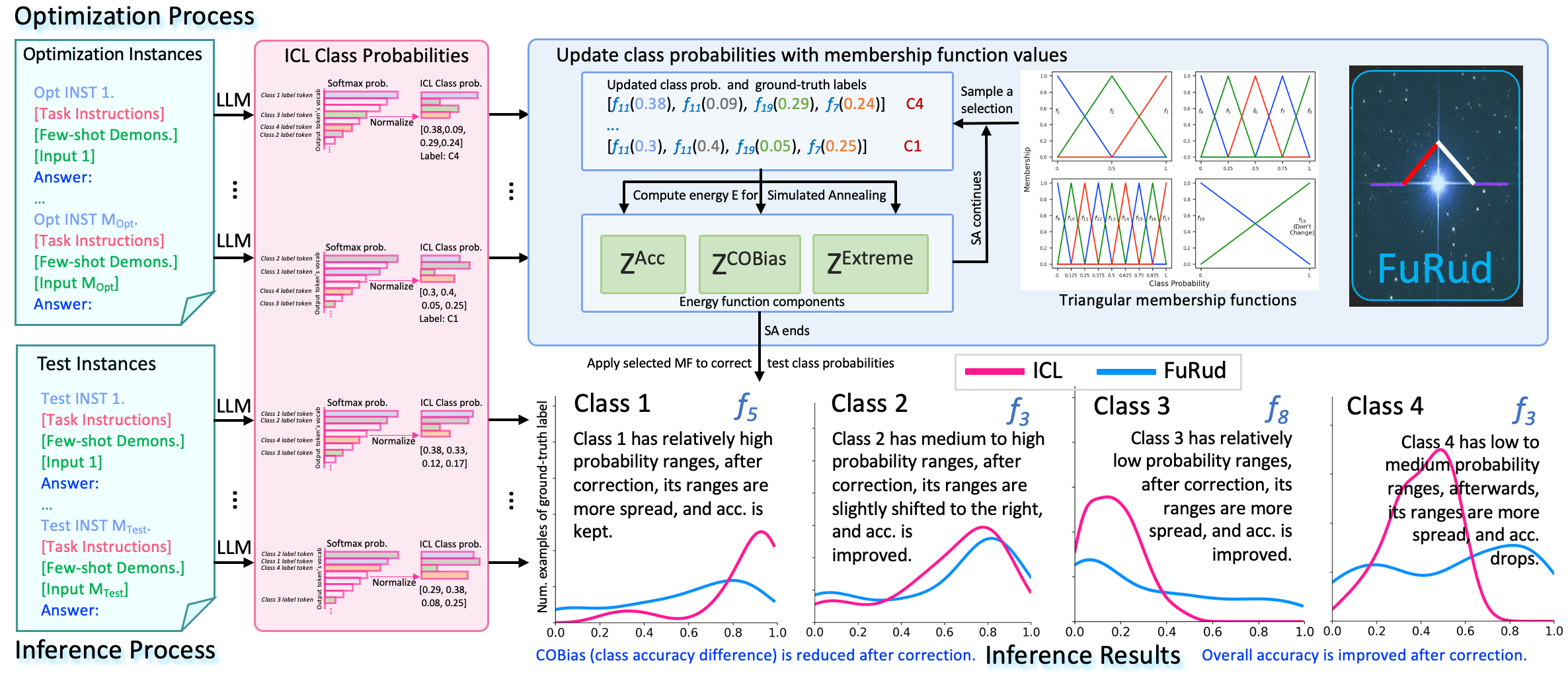}
   \caption{An overview of FuRud. ICL output probabilities across answer classes for input instances are obtained. On an optimization set, class probabilities and ground-truth labels are input to the FuRud multi-objective nonlinear integer programming model for joint learning of optimal membership functions. During inference, the optimal membership functions perform tailored corrections to class probabilities for test instances. This figure is for illustration purposes only, actual range changes and improvements are detailed in Section \ref{sec:exp}.}
     \label{fig:overview}
\end{figure*}

To this end, we propose FuRud, a \textbf{Fu}zzy \textbf{Ru}le Optimization based \textbf{D}ebiasing method, which leverages combinatorial optimization (Section \ref{sec:furud}). Optimized on a labeled set of few-shot ICL output class probabilities, each class in a downstream task selects a membership function from 19 triangular membership functions for correction, optimizing a multi-objective of COBias minimization and overall accuracy maximization. It achieves good improvements in accuracy and COBias with sample-level corrections, shown by experiments and analyses (Section \ref{sec:exp}) and discussions (Section \ref{sec:discuss}). Figure \ref{fig:overview} is an overview: an optimization set of ICL class probabilities and ground-truth answers are input to the multi-objective nonlinear integer programming model, which jointly selects optimal functions for each class. For a test instance, the learned membership functions correct the ICL class probabilities.

To highlight, the membership functions selected by FuRud enable sample-level correction and interpretability. FuRud identifies if an LLM in-context learns an accurate class probability for a given instance, namely, if \textit{Don't Change} is selected, it means the LLM has learned accurate output representations for the class; otherwise, corrections are performed on a per-sample basis. Our source code will be released upon paper publication. Our contributions are:
\begin{itemize}
    \item We propose a fuzzy rule based debiasing method (FuRud) for per-sample, per-class ICL output correction.
    \item We formulate a multi-objective nonlinear integer programming model that selects triangular membership functions for each class, to minimize class accuracy differences (COBias) and maximize overall accuracy.
    \item Across seven benchmarks, FuRud has greatly reduced COBias and improved overall accuracy. For example, it reduces the origianl ICL COBias by a relative decrease of 56\% and improves ICL accuracy by a relative increase of 21\%; it also achieves higher accuracy (avg. accuracy reaching 72.0\%) and competitive COBias (avg. COBias dropping to 17.8\%) over state-of-the-art debiasing methods.
\end{itemize}

\section{Related Work}
\label{sec:ps}
\noindent \textbf{Language Model Bias Mitigation.} At the heart of debiasing is detecting biased patterns that arise in a large language model (LLM)'s outputs. Prior work has found various prediction biases in ICL, and address the biased patterns by methods of contextual prompt engineering and output adjustment \citep{Brown2020,Schick2021a,zhao2021}. Particularly, on classification tasks, researchers have found that LLMs' outputs are sensitive to ICL formatting, such as prompt templates, demonstrations, and verbalizers \citep{min2022,holtzman2021,Schick2021b}; besides, LLMs tend to output common tokens in the pre-training data \citep{zhao2021}. These bias factors lead to majority label bias \citep{zhao2021}, COBias (pairwise class accuracy differences) \citep{cobias}, \textit{etc}, causing imbalanced per-class accuracies, and researchers address these biases by making output distribution calibrations \citep{zhao2021,dc,bc}, or by class probability re-weighting \citep{cobias}. For example, \citet{zhao2021} calibrate the output distribution with content-free/dummy test prompts. \citet{bc} calibrate the output distribution in a test-time manner, estimating a contextual correction term of each class on a batch of test examples; the proposed Batch Calibration (BC) method outperforms previous calibration methods \citep{zhao2021,dc} on a range of text classification tasks. \citet{cobias} re-weights output probabilities by a set of class-specific weight coefficients; the proposed Debiasing as Nonlinear Integer Programming method (DNIP) achieves much lower COBias with higher accuracy than the ICL baseline. Though these debiasing methods effectively adjust ICL outputs, they do not emphasize interpretable bias handling. For example, a calibration method may not explicitly explain why a class needs corrections, or users may not fathom how a re-weighting method performs the exact corrections a class need.


\section{Fuzzy Rule Optimization Based Debiasing}
\label{sec:furud}
In the fuzzy rule setting, for $N$ classes, each class selects a fuzzy set $A_{i}$, or equivalently, a membership function $f_{A_i}$, from a family of $K$ fixed fuzzy sets. We let $F=\{f_1,...,f_k,...,f_{K}\}$ denote the family of membership functions. The membership function selection problem is solved using simulated annealing. FuRud is a combinatorial optimization model, so it is performed in inference time on an optimization set of ICL output class probabilities with ground-truth labels, without LLM parameter updates. The selected membership functions are directly applied to transform test-time class probabilities.
\begin{wrapfigure}[18]{L}{0.5\textwidth}
     \begin{center}
    \includegraphics[width=0.47\textwidth]{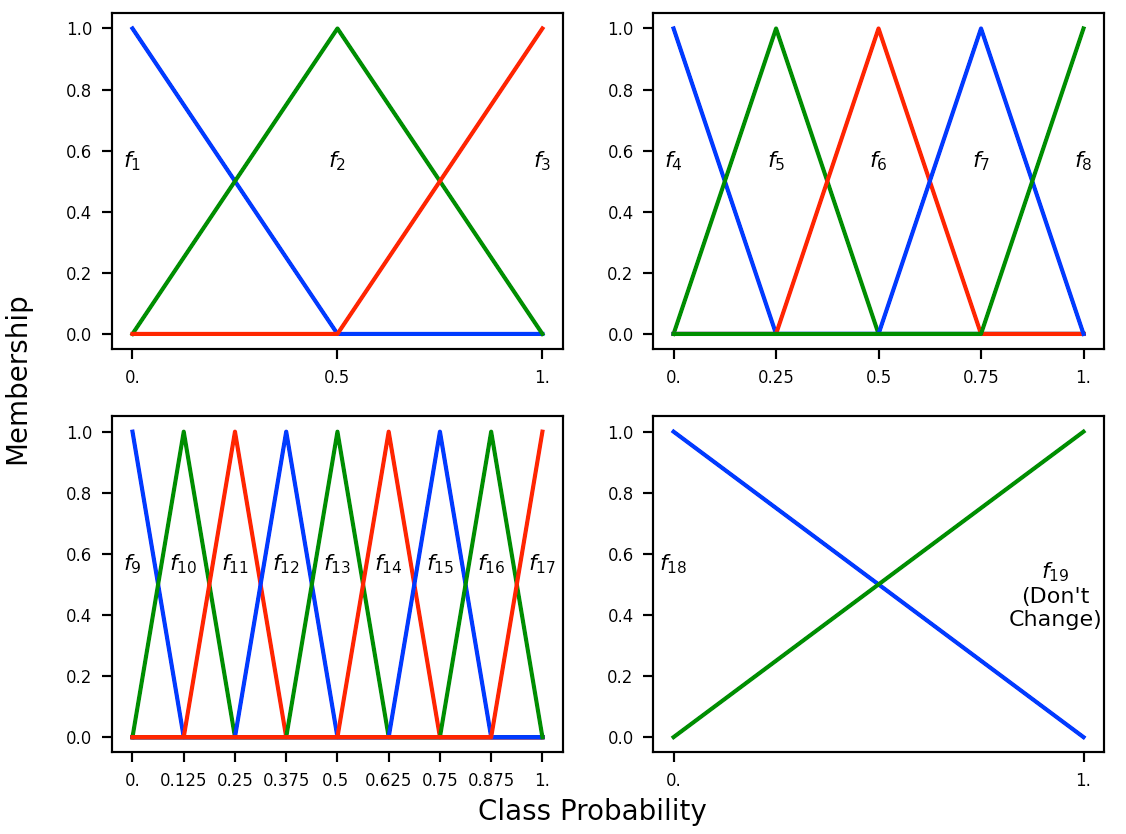}
  \end{center}
  \caption{19 triangular membership functions.}
  \label{fig:mf}
\end{wrapfigure}

\noindent \textbf{Membership Functions.}
Each class selects from 19 triangular membership functions. Triangular membership functions are popular for fuzzy rule-based classification \citep{fuzzy2005}, as the speed of changes is easily controlled by the slope, and the linearity is computationally efficient. Without knowing the expected fuzzy partitions in downstream datasets, we employ four fuzzy partitions, resulting in 19 triangular membership functions of different granularities, shown in Figure \ref{fig:mf}, including the \textit{Don't Change} membership function - the identity function. Other membership functions represent a sharp or smooth transformation of the input value. More details are provided in Appendix \ref{appdix:mfnames}. The general form of a triangular membership function $f_k(\cdot)$ can be written as:
\begin{equation}
  f_k(p_{mi}; a_k, b_k, c_k) = \left \{
  \begin{aligned}
    &0, && \text{if}\ p_{mi} \le a_k\\
    &\frac{p_{mi}-a_k}{b_k-a_k}, && a_k \le p_{mi} \le b_k\\
    &\frac{c_k-p_{mi}}{c_k-b_k}, && b_k \le p_{mi} \le c_k\\
    &0, && \text{otherwise}
  \end{aligned} \right.
  \label{eq:2}
\end{equation} 
where $a_k, b_k, c_k$ are the left endpoint, the input value where the peak is reached, and the right endpoint of $f_k$. For example, for $f_{11}$, the $a_k, b_k, c_k$ values are 0.125, 0.25, 0.375 respectively.

The updated probability $p'_{mi}$ is computed by:
\begin{equation}
  p'_{mi} = \left \{
  \begin{aligned}
    &p_{mi}, && \text{if $\textstyle \sum_{i=1}^{N} p'_{mi} =0$} \\
 \sum\limits_{k} &f_k(p_{mi})\mathbbm{1}(\kappa_i=k), && \text{otherwise}
  \end{aligned} \right.
  \label{eq:p_trans}
\end{equation} 
where $\kappa_i$ is the integer selection variable for class $i$. $\mathbbm{1}(\cdot)$ evaluates to 1 if the condition inside is satisfied, otherwise 0. Furthermore, in case $p'_{mi}=0$ for all classes, we reset each to be its original probability in $\boldsymbol{p}_m$. Therefore, $\hat{y}_m=\argmax_i p'_{mi}$.

\noindent \textbf{Multi-Objective Programming and Energy Function.}
Let $\boldsymbol{\kappa}=(\kappa_1,\dots,\kappa_N)$ be the integer selection variables for classes $1,...,N$, where $\kappa_i$ is chosen from the given set of membership functions, and $\kappa_i = k$ means $f_k$ is chosen. Our goal is to learn $\boldsymbol{\kappa}$ that improve ICL classifications under two main evaluation metrics, accuracy and COBias \citep{cobias}, i.e., our multi-objective spans across lowering COBias and increasing accuracy.

Crucially, we balance class accuracy by explicitly modeling COBias on the optimization set. The first objective is:
\begin{equation}
    \min Z\textsuperscript{COBias}=\frac{1}{\textsubscript{N}C_2} \sum\nolimits_{i=1}^{N-1} \sum_{j=i+1}^{N} \bigl| \text{Acc}_i- \text{Acc}_j \bigr|
    \label{eq:4}
\end{equation}
where $\textsubscript{N}C_2=N(N-1)/2$, $\text{Acc}_i$ is the accuracy score of class $i$ on the optimization set.

The second objective improves overall accuracy:
\begin{equation}
    \max Z\textsuperscript{Acc}=\frac{1}{|S\textsuperscript{Opt}|} \sum\nolimits_{m \in S\textsuperscript{Opt}} \mathbbm{1}\{\hat{y}_m=y_m\}
    \label{eq:5}
\end{equation}
where S\textsuperscript{Opt} is the indices of optimization instances.

To further handle extreme cases of low class accuracies, we penalize classes that fail to reach an accuracy threshold, and minimize the loss between the threshold and per-class accuracy (cut off at 0). The third objective is:
\begin{equation}
    \min Z\textsuperscript{Extreme}= \sum\nolimits_{i=1}^{N} \max \{0, \lambda-\text{Acc}_i\}
    \label{eq:6}
\end{equation}
where $\lambda$ is a fixed threshold value.

The above objective functions are a mix of minimization and maximization, and the resulted multi-objective programming model requires integer variables. Each of them alone corresponds to an integer programming problem, which is NP-complete \citep{npcomplete}. Classic solutions for integer programming use operational research techniques, such as Branch-and-Bound, often used for linear integer programming problems. It could be difficult for such methods to handle nonlinear integer programming models which contain non-differentiable functions. Consequently, a series of metaheuristic algorithms have emerged, such as Simulated Annealing (SA), and each metaheuristic has their own strengths and limitations. We use one of the metaheuristics, SA, to tackle the proposed mathematical model. The SA implementation follows \citep{cobias}. Since it is difficult to solve each one as an individual optimization problem and force an optimal solution, our strategy is instead to compute a weighted sum of $1-Z\textsuperscript{Acc}, Z\textsuperscript{COBias}, Z\textsuperscript{Extreme}$ as a single energy function $E$ to be optimized. Hence, the multi-objectives are combined into a total minimization objective:
\begin{equation}
    \min_{\kappa} E(\kappa;\lambda, \boldsymbol{p}')
    \label{eq:energy}
\end{equation}
where $E(\kappa;\lambda, \boldsymbol{p}')=\omega + \textstyle \sum_{h \in S\textsuperscript{Obj}} \gamma^h Z^h$, $S\textsuperscript{Obj}$ is the penalty (objective) functions, and $\omega, \gamma^h$s are penalty parameters. SA optimizes on $E$ to obtain an optimal selection of membership functions.

In summary, Eq. \ref{eq:4} targets minimizing COBias, Eq. \ref{eq:5} targets maximizing overall accuracy, and Eq. \ref{eq:6} targets maximizing per-class accuracy, which enforces it to meet a threshold; Eq. \ref{eq:energy} combines the three objectives as a multi-objective function. Details on how Eq. \ref{eq:energy} is optimized are described in experimental setups (Section \ref{subsec:exp_setup}).

\section{Experiments}
\label{sec:exp}
\subsection{Experimental Setups}
\label{subsec:exp_setup}
\noindent \textbf{Evaluation Tasks and Evaluation Metrics.}
The proposed method is evaluated on a diverse range of text classification datasets, including AGNews \citep{zhang2015}, a 4-class news topic classification; DBpedia \citep{auer2007}, a 14-class ontology classification dataset derived from Wikipedia; SST-5 \citep{socher2013}, a 5-class sentiment classification dataset; TREC \citep{voorhees2000,li2002}, a 6-class question classification dataset; RTE \citep{dagan2006}, a binary entailment recognition dataset; and two biomedical domain-specific datasets, including DDI \citep{ddi}, a 5-class drug-drug interaction relation extraction dataset; PubMedQA \citep{pubmedqa}, a 3-class biomedical question answering dataset. Each evaluation dataset is split into optimization/development/test sets. We follow \citep{cobias} to preprocess the datasets. Evaluation metrics are accuracy and COBias.

\begin{table*}[ht]
\small
\centering
\setlength\tabcolsep{8pt}
\begin{tabular}{@{}ccccccccc@{}}
\toprule
 &
  \multicolumn{4}{c}{Acc. $\uparrow$} &
  \multicolumn{4}{c}{COBias $\downarrow$} \\ \cmidrule(l){2-5} \cmidrule(l){6-9} 
\multirow{-2}{*}{Task} &
  ICL &
  BC &
  DNIP &
  FuRud &
  ICL &
  BC &
  DNIP &
  FuRud \\ \midrule
AGNews &
  79.9\textsubscript{7.0} &
  82.5\textsubscript{5.0} &
  87.9\textsubscript{0.7} &
  85.7\textsubscript{3.4} &
  28.3\textsubscript{16.1} &
  23.1\textsubscript{12.1} &
  6.3\textsubscript{0.6} &
  6.9\textsubscript{1.6} \\
DBpedia &
  88.6\textsubscript{1.7} &
  89.1\textsubscript{1.5} &
  93.4\textsubscript{0.6} &
  92.2\textsubscript{0.4} &
  16.2\textsubscript{3.7} &
  15.4\textsubscript{3.3} &
  7.7\textsubscript{0.6} &
  9.2\textsubscript{0.6} \\
SST-5 &
  44.9\textsubscript{4.3} &
  47.6\textsubscript{2.3} &
  48.3\textsubscript{1.9} &
  48.8\textsubscript{3.8} &
  53.1\textsubscript{5.0} &
  49.8\textsubscript{10.7} &
  18.7\textsubscript{10.1} &
  22.2\textsubscript{8.4} \\
TREC &
  68.5\textsubscript{10.8} &
  72.9\textsubscript{4.4} &
  77.1\textsubscript{2.0} &
  77.3\textsubscript{3.9} &
  35.9\textsubscript{6.5} &
  31.9\textsubscript{5.1} &
  14.2\textsubscript{1.3} &
  18.5\textsubscript{1.4} \\
RTE &
  71.5\textsubscript{2.2} &
  76.1\textsubscript{0.6} &
  74.3\textsubscript{0.8} &
  74.5\textsubscript{1.8} &
  43.4\textsubscript{7.0} &
  16.4\textsubscript{1.9} &
  4.3\textsubscript{3.3} &
  7.1\textsubscript{5.0} \\
DDI &
  7.2\textsubscript{0.9} &
  14.4\textsubscript{2.5} &
  40.4\textsubscript{6.0} &
  69.3\textsubscript{6.3} &
  45.6\textsubscript{5.9} &
  32.6\textsubscript{7.6} &
  7.5\textsubscript{3.2} &
  36.8\textsubscript{4.6} \\
PubMedaQA &
  55.1\textsubscript{2.9} &
  55.5\textsubscript{1.3} &
  63.1\textsubscript{14.0} &
  55.9\textsubscript{5.4} &
  61.2\textsubscript{1.9} &
  26.2\textsubscript{3.2} &
  41.1\textsubscript{29.6} &
  24.0\textsubscript{8.4} \\ \midrule
Avg. &
  \cellcolor[HTML]{F8E4EF}59.4 &
  \cellcolor[HTML]{F1BFD8}62.6 &
  \cellcolor[HTML]{D580AA}69.2 &
  \cellcolor[HTML]{C06895}{\color[HTML]{000000} \textbf{72.0}} &
  \cellcolor[HTML]{D4D1E2}40.5 &
  \cellcolor[HTML]{9E96B9}27.9 &
  \cellcolor[HTML]{493B7F}{\color[HTML]{FFFFFF} \textbf{14.3}} &
  \cellcolor[HTML]{5C47B3}{\color[HTML]{FFFFFF} 17.8} \\ \bottomrule
\end{tabular}
\caption{Test accuracy and COBias (\%); average scores over three runs are reported.}
\label{tab:comparison}
\end{table*}

\begin{table*}[ht]
\resizebox{\textwidth}{!}{
\begin{tabular}{@{}lcccccccc@{}}
\toprule
\textbf{\begin{tabular}[c]{@{}l@{}}Dataset,\\ Classes\end{tabular}} & \textbf{\begin{tabular}[c]{@{}c@{}}Test Sentence\\ (w/o prompt)\end{tabular}} & \textbf{\begin{tabular}[c]{@{}c@{}}Test\\ Label\end{tabular}} & \textbf{\begin{tabular}[c]{@{}c@{}}ICL Class\\ Prob.\end{tabular}} & \textbf{\begin{tabular}[c]{@{}c@{}}ICL\\ Prediction\end{tabular}} & \textbf{\begin{tabular}[c]{@{}c@{}}Membership\\ Function\end{tabular}} & \textbf{\begin{tabular}[c]{@{}c@{}}Corrected\\ Class Prob.\end{tabular}} & \textbf{\begin{tabular}[c]{@{}c@{}}Corrected\\ Prediction\end{tabular}} & \textbf{Interpretations} \\ \midrule
\rowcolor[HTML]{EFEFEF} 
\begin{tabular}[c]{@{}l@{}}AGNews\\ World, Sports,\\ Business, Tech\end{tabular} & \begin{tabular}[c]{@{}c@{}}US unemployment claims slip but\\ picture still murky, NEW YORK\\ Fewer Americans lined up to claim first-time\\ jobless benefits last week but analysts said\\ the modest decline said very little about\\ the current state of the labour market.\end{tabular} & Business & \begin{tabular}[c]{@{}c@{}}{[}0.42, 0.01, \\ 0.32, 0.25{]}\end{tabular} & World & \begin{tabular}[c]{@{}c@{}}$f_7, f_{16}$, \\ $f_{11}, f_7$\end{tabular} & \begin{tabular}[c]{@{}c@{}}{[}0, 0,  \\ 0.47, 0{]}\end{tabular} & Business & \begin{tabular}[c]{@{}c@{}}By FuRud, all classes need\\ corrections. For this test instance,\\ original  ICL wrongly predicts\\ Busi. as World. After correction,\\ probability of class Busi. becomes\\ highest, leading to the right prediction.\end{tabular} \\
\begin{tabular}[c]{@{}l@{}}DBpedia\\ Company, School, Artist,  Athlete, \\ Politician, Transportation, Building,\\ Nature, Village, Animal, Plant,\\ Album, Film, Book\end{tabular} & \begin{tabular}[c]{@{}c@{}}Floyd Thomas Christian Sr.\\ (December 18 1914 – May 11 1998)\\ was Florida Commissioner of Education\\ from 1965 to 1973. Christian was born in\\ Bessemer Alabama. He moved to Pinellas\\ County with his family in 1927...\end{tabular} & Politician & \begin{tabular}[c]{@{}c@{}}{[}0, 0.16, 0.08, 0.64,\\ 0.12, 0, 0, 0,\\ 0, 0, 0, 0,\\ 0, 0{]}\end{tabular} & Athlete & \begin{tabular}[c]{@{}c@{}}$f_3, f_{16}, f_7, f_8$,\\ DC, $f_2, f_2, f_2$,\\ $f_{16}, f_2, f_{16}$, DC,\\ DC, DC\end{tabular} & \begin{tabular}[c]{@{}c@{}}{[}0, 0, 0, 0,\\ 0.12, 0, 0, 0,\\ 0, 0, 0, 0,\\ 0, 0{]}\end{tabular} & Politician & \begin{tabular}[c]{@{}c@{}}By FuRud, 4 out of 14 classes\\ apply Don't Change (Their ICL\\ probability is relatively accurate.),\\ including class Politician. Though ICL\\probability of the actual class Politic. is 0.12 and\\ unchanged after correction, classes like\\ Ath.'s probability is corrected to 0 by f8,\\ leading to the right prediction.\end{tabular} \\ \bottomrule
\end{tabular}
}
\caption{Examples of sample-level corrections and explanations.}
\label{tab:example}
\end{table*}

\begin{figure*}[ht]
    \centering
    \includegraphics[width=\linewidth]{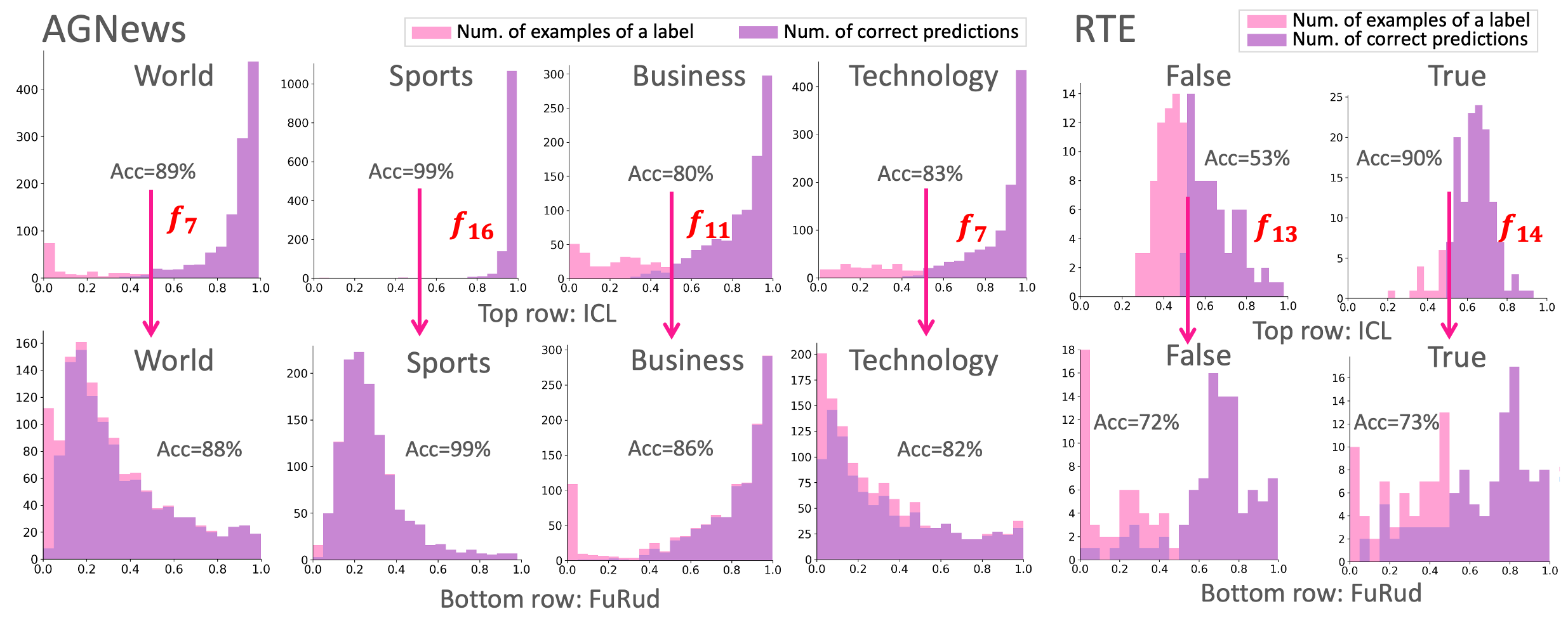}
   \caption{Class probabilities before and after applying corrections. For each task, we report results of the seed 1 run out of 3 runs. There was a stark ICL accuracy difference of 37\% between \textit{True} and \textit{False} on RTE. FuRud addresses it by amplifying the medium range of \textit{False} and simultaneously reducing the relatively high range of \textit{True}.}
   \label{fig:inter_vis}
\end{figure*}

\begin{figure*}[ht]
    \centering
    \includegraphics[width=\linewidth]{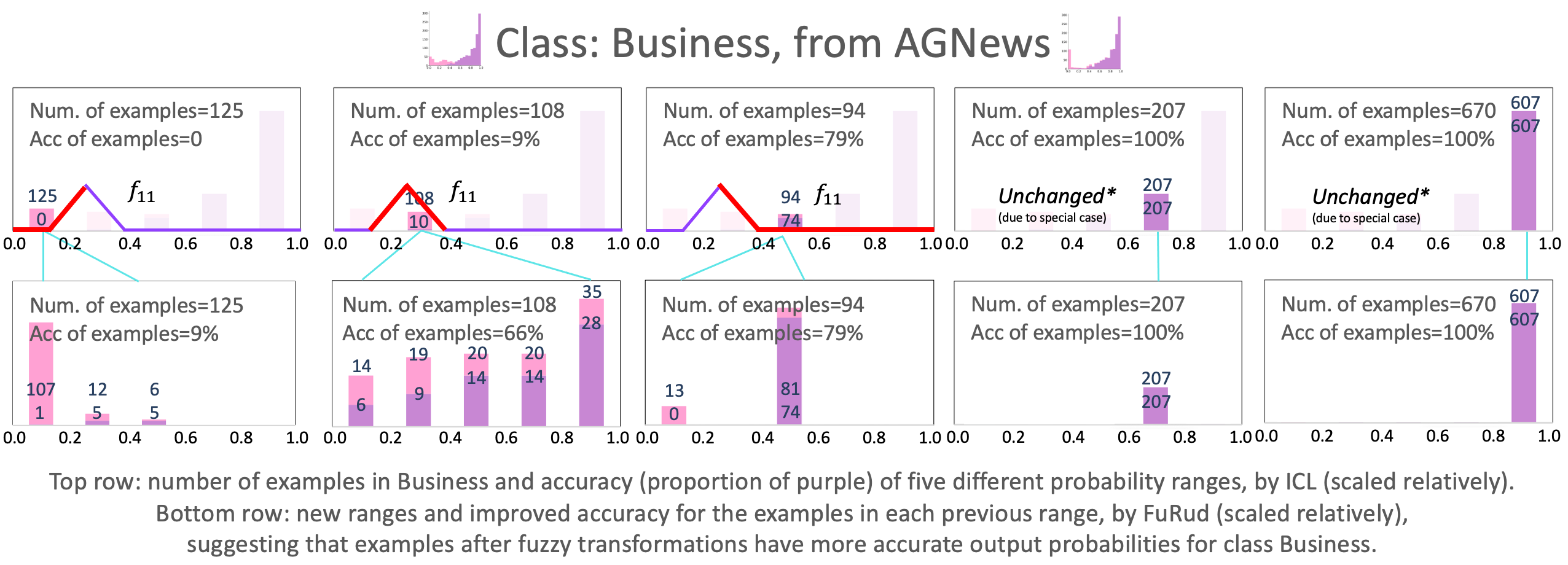}
   \caption{Quantitative evaluation on the ratio of instances that benefit from the correction, exemplified by class \textit{Business} of AGNews. The difference of ``Acc of examples'' between bottom and top subfigures represents the ratio. The red color highlights the activated pieces of the membership function for range-specific correction.}
   \label{fig:transform}
\end{figure*}

\noindent \textbf{FuRud Setups.}
The 19 triangular membership functions in Figure \ref{fig:mf} form the base of selections for FuRud. We take the full or a subset of training instances from a downstream dataset to perform FuRud optimization. We prompt Llama-2-13B in 1-shot manner to obtain softmax probabilities at the output token over the entire vocabulary, which is then normalized over the classes. These ICL class probabilities and ground-truth labels are used to form the optimization set. The energy function we used is a special form of Equation \ref{eq:energy} with $\omega=1, \gamma\textsuperscript{Acc}=-1, \gamma\textsuperscript{COBias}=\alpha, \gamma\textsuperscript{Extreme}=\beta$. That is, the final multi-objective optimization function is $min_{\kappa} Z=1-Z\textsuperscript{Acc}+\alpha Z\textsuperscript{COBias} +\beta Z\textsuperscript{Extreme}$, where we learn $\kappa_i$ for class $i=1,\dots,N$ on the optimization set. Each $\kappa_i$ is selected from the given set of membership functions, and $\kappa_i = k$ denotes that membership function $f_k$ is selected. At inference time, let $p = (p_1,\dots, p_i,\dots,p_N)$ be the ICL class probabilities of a test instance, then these probabilities are transformed by the learned membership functions, according to Eq. \ref{eq:p_trans}. The final corrected prediction is $\hat{y}= \argmax_i f_{\kappa_i} (p_i)$. 

The FuRud model $Z$ is solved using simulated annealing (SA). The core step of SA is to sample a new solution $\kappa = (\kappa_1,\dots, \kappa_N)$, e.g., (16, $\dots$, 8), which is evaluated against $Z$. If the new $Z$ is smaller, FuRud accepts the new solution; otherwise, it accepts the new solution with an acceptance probability $exp(-\Delta Z/T)$, where $T$ is the temperature at the step. The values of $\alpha, \beta$ are tuned on the development set. Since we do not know an estimate for the expected threshold value $\lambda$ in downstream tasks, we set it to 0.5 for simplicity. Prompting is done on a 80G A100 GPU. The SA algorithm executes on an AMD EPYC 7742 CPU in minutes.

We compare FuRud with the ICL baseline and two state-of-the-art ICL debiasing methods, including DNIP \citep{cobias} and BC \citep{bc}. For fair comparisons, for each dataset, we prompt with three different 1-shot demonstrations and obtain three sets of initial probabilities. The demonstration is randomly sampled from optimization examples. The average test accuracy and COBias over three runs are reported.

\subsection{Main Results}
Table \ref{tab:comparison} shows the test accuracy and COBias of ICL, BC, DNIP, and FuRuD. Comparing FuRud to the ICL baseline, the average relative accuracy increase is 21\%, and the average relative COBias reduction is 56\%. The average test accuracy of FuRud over seven benchmarks is 72\%, which outperforms the accuracy of BC and DNIP; the average test COBias of FuRud is 17.8\%, which is comparable to DNIP with obtains the lowest COBias (14.3\% ) among the methods compared. It is noted that FuRud uses the full optimization set to make a fair comparison to DNIP. However, FuRud can also work in a few-shot optimization manner, as discussed in Appendix \ref{appdix:fs}. On top of that, FuRud provides per-sample, per-class interpretability, analyzed as follows.

\subsection{Interpretability Analysis}
Across seven tasks, AGNews is the only task that \textit{Don't Change} was not applied to any class over any of its three runs with different initial ICL probabilities; RTE, DDI, and PubMedQA applied \textit{DC} to a single class at most; TREC and SST-5 applied \textit{DC} to a two classes at most; DBpedia, with 14 classes, at most applied \textit{DC} to 4 classes, showing that most ICL output classes need correction.

In addition, Figure \ref{fig:inter_vis} visualizes range-specific probability changes after applying membership function corrections on AGNews and RTE, demonstrating that the membership functions selected by FuRud effectively transform different probability ranges of each output class to improve or at least maintain class accuracy, while making class accuracies more balanced. Moreover, the worst-performing class by ICL in either task is significantly improved by FuRud, due to that lower and medium probability ranges of the worst-performing class gets amplified and the higher ranges of other classes gets slightly reduced. Results on other datasets are similar. Table \ref{tab:example} exemplifies how per-sample, per-class corrections are interpreted.

We further quantitatively evaluate the ratio of instances that benefit from the correction. Figure \ref{fig:transform} shows the accuracy of range-specific instances within class \textit{Business} of AGNews. This class has 1,204 test instances, which are divided into 5 groups ranging from $[0.0, 0.2]$ to $[0.8, 1.0]$ based on their initial ICL output probability in \textit{Business}. For example, 108 instances have ICL \textit{Business} probabilities in $[0.2, 0.4]$ and only 9\% of these instances are correctly predicted. This group of instances is effectively corrected by membership function $f_{11}$ and synergetic corrections in other classes, reaching 66\% accuracy after correction, i.e., 57\% (66\%-9\%) more instances in this group obtain the right predictions after corrections. 

\section{Discussion}
\label{sec:discuss}
\subsection{FuRud on Letter Based ICL Outputs}
FuRud greatly improves highly skewed letter based ICL Outputs. In details, we experiment with the letter answer prompts, which is widely used in classification and reasoning tasks. Letter options could lead to more shallow pattern matching problems than using label token as answer options. Using this prompt, the model outputs a single letter choice of ``A'', ``B'', \textit{etc.} corresponding to a class label, which often results in highly skewed outputs, because LLMs have a tendency to select a certain letter option regardless of the content \citep{letter}. We find similar issues when evaluating seven datasets using letter options. For example, on AGNews, the LLM biases to predict ``B'' (\textit{Sports}), leading to an average of 99\% accuracy in \textit{Sports} and 12\% accuracy in \textit{Business}. FuRud improves accuracy by an relative 44\% and achieves a significant COBias reduction of a relative 54\% over ICL, as shown in Table \ref{tab:letter} (averaged over seven datasets).
\begin{wraptable}{r}{4.5cm}
\small
\centering
\resizebox{0.28\textwidth}{!}{%
\begin{tabular}{ccc}
\hline
Method         & Acc.                     & COBias                   \\ \hline
\rowcolor[HTML]{ECF4FF} 
ICL (letter)   & 36.9\textsubscript{13.6} & 47.2\textsubscript{15.6} \\
\rowcolor[HTML]{CBCEFB} 
FuRud (letter) & 53.1\textsubscript{10.5} & 21.6\textsubscript{8.2}  \\ \hline
\end{tabular}
}
\caption{Letter based results.}
\label{tab:letter}
 \end{wraptable}
Besides the tabled results, on AGNews, overall test accuracy improves from 45\% to 66\%, COBias reduces from 54\% to 10\%. Class accuracy changes are: \textit{World}, 40\% $\rightarrow$ 69\%; \textit{Sports}, 99\% $\rightarrow$ 70\%; \textit{Business}, 12\% $\rightarrow$ 66\%; \textit{Technology}, 27\% $\rightarrow$ 59\%. These results suggest that FuRud can debias ICL class probabilities no matter if an input prompt leads to spurious label matching results.

\subsection{Membership Function Granularities}
We experiment with different combinations of the four fuzzy partitions in Figure \ref{fig:mf} and show that membership granularities lead to accuracy-COBias tradeoffs. In details, we conduct five ablations based on the four partitions characterized by different slopes $\pm 1, \pm 2, \pm 4, \pm 8$, where a bigger slope indicates higher granularities. The $\pm 1$ partition is the DC partition ($f_{18}$, $f_{19}$). Since it plays a vital role in maintaining some classes, we keep it in all five combinations, including DC alone, DC and each of the rest partitions, and mixed partitions (all four partitions). The average scores across seven datasets are reported; for each dataset, average accuracy and COBias over three runs with different demonstrations is taken.
\begin{wrapfigure}[13]{R}{0.3\textwidth}
  \begin{center}
    \includegraphics[width=0.3\textwidth]{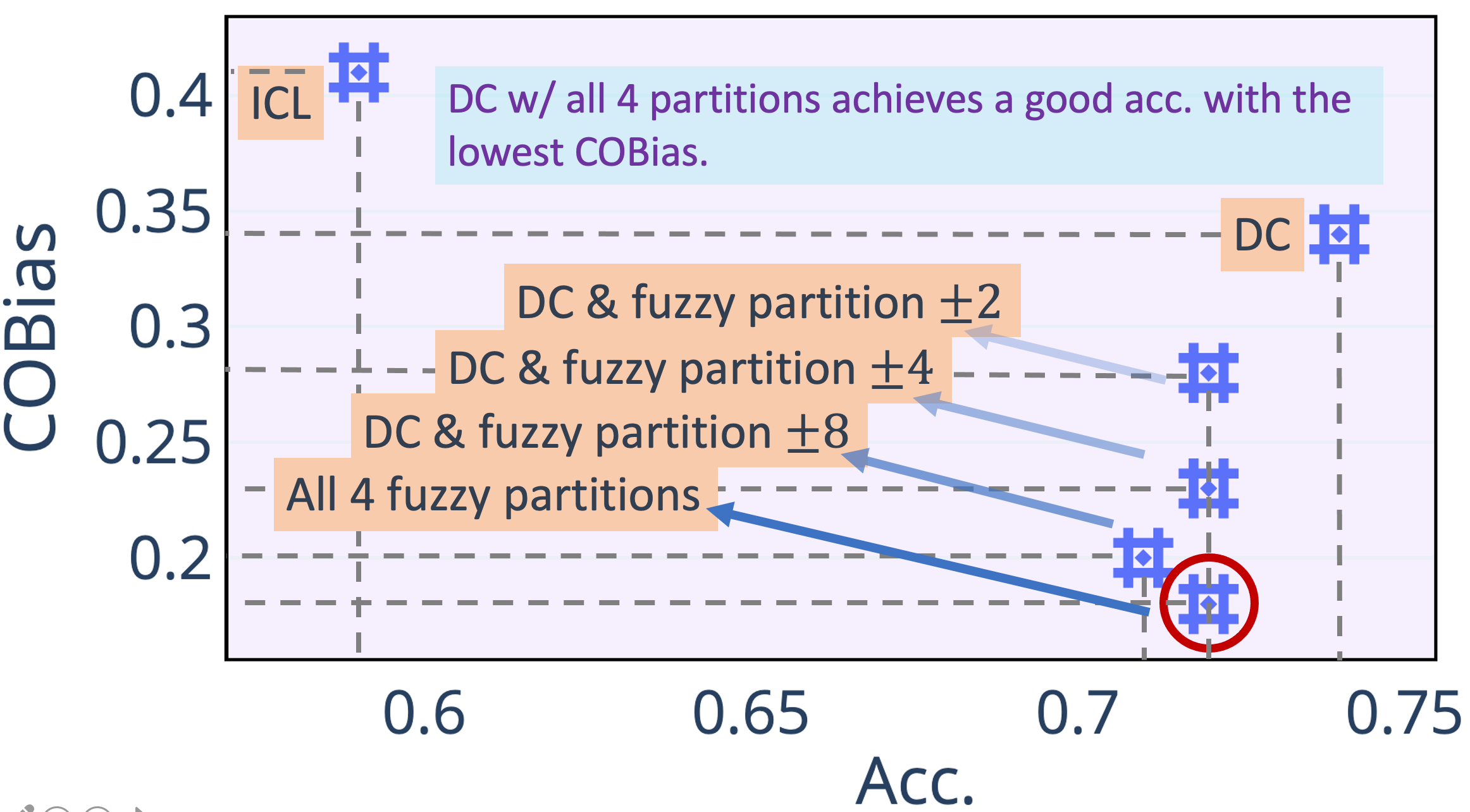}
  \end{center}
  \caption{Accuracy-COBias tradeoff of fuzzy partitions.}
  \label{fig:granu}
\end{wrapfigure}

As shown by Figure \ref{fig:granu}, while COBias greatly reduces with higher membership granularities, overall accuracy slightly decreases. Therefore, although it is tempting to include more fine-grained membership functions to reduce COBias, don't forget the accuracy-COBias tradeoff. The best accuracy and COBias is achieved with mixed partitions. Moreover, although he DC partition alone can obtain 15\% higher accuracy than ICL accuracy, but the improvement mainly comes from a single task (DDI). In addition, we added a side analysis with partition $\pm 8$ alone, while achieving similar accuracies, the COBias is 6\% higher than using DC and partition $\pm 8$ together, suggesting that the \textit{Don’t Change} function is essentially needed when using mixed partitions to attain both good COBias and overall accuracy.

\subsection{Additional Analyses}
\label{subsec:dnip_inter}
\noindent \textbf{More Models.} On varied relatively small LLMs, FuRud consistently improves their performances, showcased by additional results on Llama-2-7B and GPT-2-XL in Appendix \ref{appdix:morellms}.

\noindent \textbf{More ICL Strategies.} FuRud significantly improves accuracy and COBias for a more sophisticated prompting case, where the demonstration in the prompt uses $N$-shot examples taken from each class, as detailed in Appendix \ref{appdix:moreicl}.

\noindent \textbf{Computational Costs.} The computational cost of FuRud is low. Execution time of the optimization program ranges rom several minutes to around 30 minutes, depending on the number of classes, optimization set sizes, and etc.

\noindent \textbf{Interpretability Comparisons: FuRud vs DNIP.} DNIP does not capture sample-level nuances needed in the correction as FuRud does, not suitable for classes that need fine-grained sample-level correction. The membership functions overcome this issue, explaining how each class in a particular instance should be corrected, and this is the main innovation of our paper. 

\noindent \textbf{Using traditional fuzzy classification systems.} Training involves extensive computations as it requires maintaining multiple candidate rules for debiasing, e.g., ``$R_q$: If class 1 probability is $A_{q1}$ and $\dots$ and class $N$ probability is $A_{qN}$, then predict $Y_q$'', and test time for a winning rule grows with the number of candidates. To obtain high accuracy, a huge number of rules may be employed, making the system inefficient. In contrast, FuRud is cost-effective, and implicitly reflects the many rules employed in a traditional system. For more discussions, please refer to Appendix \ref{appdix:morediscuss}.

\section{Conclusion}
We present FuRud, a post-hoc debiasing method for sample-level ICL output correction that effectively enhances overall accuracy and balanced accuracy across multiple classes, leveraging combinatorial optimization to select optimal fuzzy membership functions for interpretable, range-specific corrections on the original ICL class probabilities. On a diverse set of text classification benchmarks, FuRud greatly reduces COBias while enhancing overall accuracy over ICL results, outperforming state-of-the-art debiasing methods. 

\section*{Limitations}
There is a possibility that certain classes in a task may not need as much fine-grained sample-level correction as other classes. Those classes may apply class-level correction to any instance, e.g., weight coefficients, to obtain sufficient COBias reduction while being interpretable at the broader level. As such, an organic integration of both broad and fine-grained corrections should be studied in the future. In addition, class accuracy imbalances widely exist in variants of small (e.g., Llama-2-13B) and larger (e.g., ChatGPT) LLMs \citep{cobias}, how much FuRud could balance larger LLMs could be quantitatively analyzed to exemplify more practical usages of the method. This paper primarily focuses on a single small model for its representativeness as a widely applied LLM.

\bibliography{output}{}
\bibliographystyle{unsrtnat}

\newpage
\appendix

\section{Details on Membership Functions}
\label{appdix:mfnames}

Table \ref{tab:mfnames} lists the details about the membership functions used in this work.

\begin{table*}[ht]
\centering
\scalebox{0.7}{
\begin{tabular}{@{}lcccl@{}}
\toprule
\textbf{Function} & \textbf{Parameters} & \textbf{Name}    & \textbf{Short Form} & \multicolumn{1}{c}{\textbf{Meaning}}                                                                                               \\ \midrule
\rowcolor[HTML]{ECF4FF} 
$f_1$             & 0, 0, 0.5           & Low-2            & L-2                 & \begin{tabular}[c]{@{}l@{}}Low-range transformation,\\ smooth change with slope $-2$, peak at 0\end{tabular}                       \\
$f_2$             & 0, 0.5, 1           & Medium-2         & M-2                 & \begin{tabular}[c]{@{}l@{}}Medium-range transformation,\\ smooth change with slope $\pm 2$, peak at 0.5\end{tabular}               \\
\rowcolor[HTML]{ECF4FF} 
$f_3$             & 0.5, 1, 1           & High-2           & H-2                 & \begin{tabular}[c]{@{}l@{}}High-range transformation,\\ smooth change with slope $2$, peak at 1\end{tabular}                       \\ \midrule
$f_4$             & 0, 0, 0.25          & Low-4            & L-4                 & \begin{tabular}[c]{@{}l@{}}Low-range transformation,\\ sharp change with slope $-4$, peak at 0\end{tabular}                        \\
\rowcolor[HTML]{ECF4FF} 
$f_5$             & 0, 0.25, 0.5        & Medium Low-4     & ML-4                & \begin{tabular}[c]{@{}l@{}}Low-to-medium-range transformation,\\ sharp change with slope $\pm 4$, peak at 0.25\end{tabular}        \\
$f_6$             & 0.25, 0.5, 0.75     & Medium-4         & M-4                 & \begin{tabular}[c]{@{}l@{}}Medium-range transformation,\\ sharp change with slope $\pm 4$, peak at 0.5\end{tabular}                \\
\rowcolor[HTML]{ECF4FF} 
$f_7$             & 0.5, 0.75, 1        & Medium High-4    & MH-4                & \begin{tabular}[c]{@{}l@{}}Medium-to-high-range transformation,\\ sharp change with slope $\pm 4$, peak at 0.75\end{tabular}       \\
$f_8$             & 0.75, 1, 1          & High-4           & H-4                 & \begin{tabular}[c]{@{}l@{}}High-range transformation,\\ sharp change with slope $4$, peak at 1\end{tabular}                        \\ \midrule
\rowcolor[HTML]{ECF4FF} 
$f_9$             & 0, 0, 0.125         & Very Very Low-8  & VVL-8               & \begin{tabular}[c]{@{}l@{}}Very-very-low-range transformation,\\ very sharp change with slope $-8$, peak at 0\end{tabular}         \\
$f_{10}$          & 0, 0.125, 0.25      & Very Low-8       & VL-8                & \begin{tabular}[c]{@{}l@{}}Very-low-range transformation,\\ very sharp change with slope $\pm 8$, peak at 0.125\end{tabular}       \\
\rowcolor[HTML]{ECF4FF} 
$f_{11}$          & 0.125, 0.25, 0.375  & Low-8            & L-8                 & \begin{tabular}[c]{@{}l@{}}Low-range transformation,\\ very sharp change with slope $\pm 8$, peak at 0.25\end{tabular}             \\
$f_{12}$          & 0.25, 0.375, 0.5    & Medium Low-8     & ML-8                & \begin{tabular}[c]{@{}l@{}}Low-to-medium-range transformation,\\ very sharp change with slope $\pm 8$, peak at 0.375\end{tabular}  \\
\rowcolor[HTML]{ECF4FF} 
$f_{13}$          & 0.375, 0.5, 0.625   & Medium-8         & M-8                 & \begin{tabular}[c]{@{}l@{}}Medium-range transformation,\\ very sharp change with slope $\pm 8$, peak at 0.5\end{tabular}           \\
$f_{14}$          & 0.5, 0.625, 0.75    & Medium High-8    & MH-8                & \begin{tabular}[c]{@{}l@{}}Medium-to-high-range transformation,\\ very sharp change with slope $\pm 8$, peak at 0.625\end{tabular} \\
\rowcolor[HTML]{ECF4FF} 
$f_{15}$          & 0.625, 0.75, 0.875  & High-8           & H-8                 & \begin{tabular}[c]{@{}l@{}}High-range transformation,\\ very sharp change with slope $\pm 8$, peak at 0.75\end{tabular}            \\
$f_{16}$          & 0.75, 0.875, 1      & Very High-8      & VH-8                & \begin{tabular}[c]{@{}l@{}}Very-high-range transformation,\\ very sharp change with slope $\pm 8$, peak at 0.875\end{tabular}      \\
\rowcolor[HTML]{ECF4FF} 
$f_{17}$          & 0.875, 1, 1         & Very Very High-8 & VVH-8               & \begin{tabular}[c]{@{}l@{}}Very-very-high-range transformation,\\ very sharp change with slope $8$, peak at 1\end{tabular}         \\ \midrule
$f_{18}$          & 0, 0, 1             & Full-1           & F-1                 & \begin{tabular}[c]{@{}l@{}}Full-range transformation,\\ very smooth change with slope $-1$, peak at 0\end{tabular}                 \\
\rowcolor[HTML]{ECF4FF} 
$f_{19}$          & 0, 1, 1             & Don't Change     & Don't Change        & Identity function                                                                                                                  \\ \bottomrule
\end{tabular}
}
\caption{Names, parameters ($a, b, c$), short forms, and meanings for membership functions.}
\label{tab:mfnames}
\end{table*}

\section{Few-shot Optimization}
\label{appdix:fs}
\begin{wrapfigure}[17]{R}{0.7\textwidth}
  \begin{center}
    \includegraphics[width=0.7\textwidth]{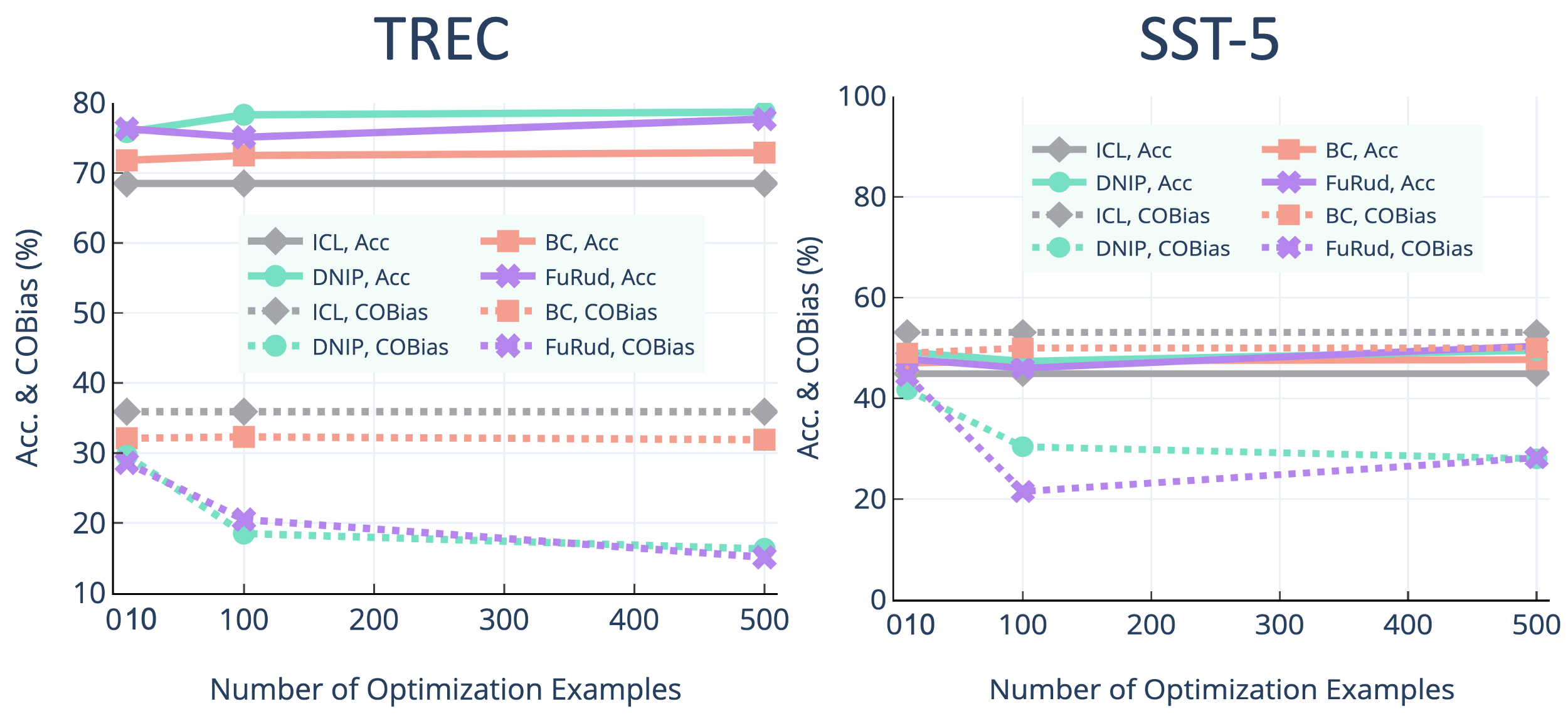}
  \end{center}
  \caption{Few-shot optimization.}
  \label{fig:fs_opt}
\end{wrapfigure}

FuRud can optimize a downstream task with as few as 10 examples. We take few-shot optimized TREC and SST-5 results for illustration. Figure \ref{fig:fs_opt} shows test accuracy and COBias of FuRud (in mint green color) when used in a few-shot optimization manner, starting with 10 few-shot examples and growing to 100 and 500 examples. TREC and SST-5 are shown to illustrate that FuRud can achieve an average of 9\% accuracy improvements with 18\% COBias reduction over the ICL baseline at 10 few-shot optimization examples.

\begin{table*}[ht]
\setlength\tabcolsep{3.7pt}
\small
\centering
\begin{tabular}{@{}cccccccccc@{}}
\toprule
{Model} &
  {Metric} &
  {AGNews} &
  {DBpedia} &
  {SST-5} &
  {TREC} &
  {RTE} &
  {DDI} &
  {PubMedQA} &
  {Avg.} \\ \midrule
\multicolumn{10}{c}{{Llama-2-7B}} \\ \midrule
{ICL} &
  {Acc} &
  {$86.4_{2.5}$} &
  {$88.9_{2.0}$} &
  {$42.1_{11.1}$} &
  {$66.7_{6.6}$} &
  {$66.3_{4.3}$} &
  {$6.7_{0.4}$} &
  {$40.3_{6.7}$} &
  {56.8} \\
{} &
  {COBias} &
  {$14.0_{6.5}$} &
  {$13.5_{2.1}$} &
  {$55.6_{1.5}$} &
  {$33.2_{10.0}$} &
  {$61.6_{10.5}$} &
  {$41.4_{1.7}$} &
  {$40.9_{16.1}$} &
  {37.2} \\ \midrule
{FuRud} &
  {Acc} &
  {$\boldsymbol{88.5_{0.5}}$} &
  {$\boldsymbol{91.5_{0.5}}$} &
  {$\boldsymbol{49.5_{0.7}}$} &
  {$\boldsymbol{73.1_{3.9}}$} &
  {$\boldsymbol{72.7_{1.0}}$} &
  {$\boldsymbol{54.4_{6.4}}$} &
  {$\boldsymbol{55.7_{7.6}}$} &
  {$\boldsymbol{69.3}$} \\
{} &
  {COBias} &
  {$\boldsymbol{7.4_{2.5}}$} &
  {$\boldsymbol{8.4_{0.6}}$} &
  {$\boldsymbol{24.0_{1.2}}$} &
  {$\boldsymbol{14.1_{1.9}}$} &
  {$\boldsymbol{4.2_{2.7}}$} &
  {$\boldsymbol{16.9_{5.0}}$} &
  {$\boldsymbol{21.8_{16.6}}$} &
  {$\boldsymbol{13.8}$} \\ \midrule
\multicolumn{10}{c}{{GPT2-XL}} \\ \midrule
{ICL} &
  {Acc} &
  {$52.1_{5.4}$} &
  {$31.8_{9.9}$} &
  {$34.9_{13.7}$} &
  {$27.4_{10.5}$} &
  {$55.4_{1.9}$} &
  {$14.5_{4.4}$} &
  {$55.2_{0.0}$} &
  {38.8} \\
{} &
  {COBias} &
  {$35.5_{11.5}$} &
  {$40.0_{3.6}$} &
  {$48.7_{5.4}$} &
  {$45.6_{8.7}$} &
  {$82.4_{24.5}$} &
  {$40.7_{5.9}$} &
  {$59.4_{12.6}$} &
  {50.3} \\
{FuRud} &
  {Acc} &
  {$\boldsymbol{69.0_{0.5}}$} &
  {$\boldsymbol{67.7_{11.8}}$} &
  {$\boldsymbol{43.4_{3.1}}$} &
  {$\boldsymbol{41.7_{2.7}}$} &
  {$\boldsymbol{51.2_{3.7}}$} &
  {$\boldsymbol{53.2_{17.0}}$} &
  {$\boldsymbol{48.4_{0.3}}$} &
  {$\boldsymbol{53.5}$} \\
{} &
  {COBias} &
  {$\boldsymbol{7.4_{2.9}}$} &
  {$\boldsymbol{23.0_{6.5}}$} &
  {$\boldsymbol{25.4_{1.4}}$} &
  {$\boldsymbol{30.2_{7.0}}$} &
  {$\boldsymbol{8.9_{3.6}}$} &
  {$\boldsymbol{23.1_{6.5}}$} &
  {$\boldsymbol{17.6_{4.6}}$} &
  {$\boldsymbol{19.4}$} \\ \bottomrule
\end{tabular}
\caption{Test accuracy and COBias Comparisons on more LLMs.}
\label{tab:morecomp}
\end{table*}

\begin{table*}[ht]
\small
\centering
\setlength\tabcolsep{2.2pt}
\begin{tabular}{@{}cccccccccc@{}}
\toprule
{\begin{tabular}[c]{@{}c@{}}Demonstration\\ Selection\end{tabular}} &
  {Metric} &
  {AGNews} &
  {DBpedia} &
  {SST-5} &
  {TREC} &
  {RTE} &
  {DDI} &
  {PubMedQA} &
  {Avg.} \\ \midrule
{k-shot ICL} &
  {Acc} &
  {$83.5_{1.5}$} &
  {$95.2_{1.2}$} &
  {$50.3_{2.3}$} &
  {$67.0_{12.7}$} &
  {$75.0_{0.8}$} &
  {$9.7_{1.0}$} &
  {$52.3_{5.3}$} &
  {61.9} \\
{} &
  {COBias} &
  {$14.9_{5.1}$} &
  {$7.0_{2.2}$} &
  {$36.3_{7.2}$} &
  {$38.2_{5.1}$} &
  {$22.5_{13.2}$} &
  {$39.7_{3.5}$} &
  {$20.9_{4.2}$} &
  {25.6} \\ \midrule
{FuRud} &
  {Acc} &
  {$\boldsymbol{88.1_{0.6}}$} &
  {$\boldsymbol{96.6_{0.4}}$} &
  {$\boldsymbol{54.3_{1.3}}$} &
  {$\boldsymbol{77.9_{6.0}}$} &
  {$\boldsymbol{75.9_{4.6}}$} &
  {$\boldsymbol{62.3_{2.1}}$} &
  {$\boldsymbol{59.2_{5.9}}$} &
  {$\boldsymbol{73.5}$} \\
{} &
  {COBias} &
  {$\boldsymbol{7.7_{2.5}}$} &
  {$\boldsymbol{4.4_{0.7}}$} &
  {$\boldsymbol{13.8_{4.1}}$} &
  {$\boldsymbol{11.6_{3.3}}$} &
  {$\boldsymbol{5.0_{1.4}}$} &
  {$\boldsymbol{27.0_{2.2}}$} &
  {$\boldsymbol{21.3_{8.7}}$} &
  {$\boldsymbol{13.0}$} \\ \bottomrule
\end{tabular}
\caption{Test accuracy and COBias under the k-shot demonstration selection strategy.}
\label{tab:kshot}
\end{table*}

\begin{figure*}[ht]
    \centering
    \small
    \includegraphics[width=\textwidth]{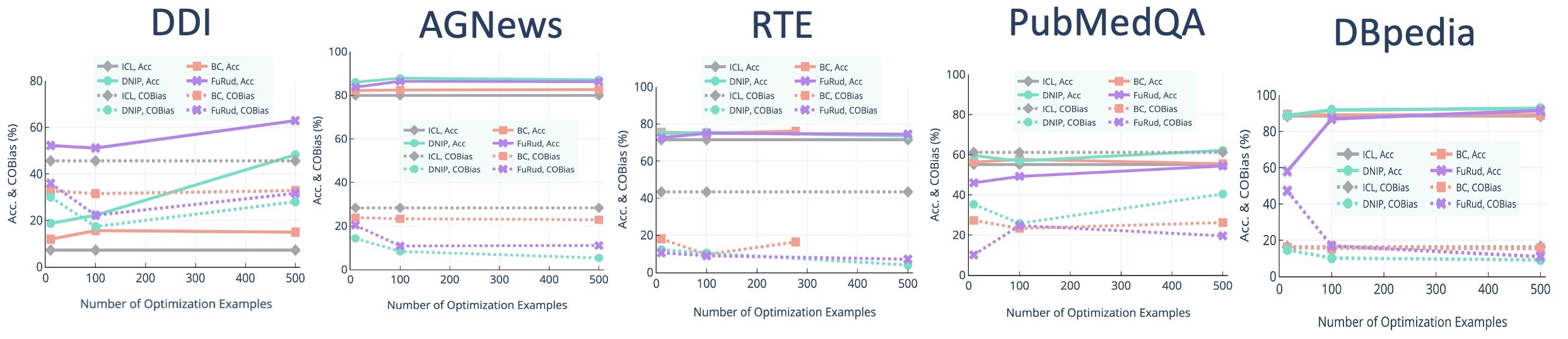}
   \caption{Additional few-shot optimization results.}
   \label{fig:appdix_opt}
\end{figure*}

At 10 examples, FuRud obtains a 11\% and 6\% relative increase in accuracy over the ICL baseline on TREC and SST-5 respectively, at the same time, it reduces COBias by a relative 20\% and 16\% on each dataset. The accruacy and COBias performances gradually improve as the number of examples increases to 500. Compared to existing methods, FuRud outperforms BC in few-shot scenarios, and performs better than (TREC) or on par (SST-5) with DNIP while being interpretable. Similar findings apply to the other five datasets, as shown in Figure \ref{fig:appdix_opt}. In short, FuRud achieves better or comparable results than DNIP, and better results than BC and the ICL baseline, while providing enhanced interpretability.

\section{FuRud's Performances on More LLMs}
\label{appdix:morellms}
We ran experiments of FuRud on two additional models, Llama-2-7B and GPT2-XL. Results are shown in Table \ref{tab:morecomp}. For example, on Llama-2-7B, FuRud improves accuracy by a relative 22\%, and reduces COBias by a relative 63\% over ICL baselines, demonstrating that FuRud gains consistent performance improvements on various models. Indeed, our current evaluations are focused on relatively small LLMs, but our approach can also work for larger models, as long as class probabilities are available and the imbalanced per-class accuracy issue exists.

\section{FuRud's Performances under More ICL Demonstration Selection Strategies}
\label{appdix:moreicl}

We additionally prompt Llama-2-13B with the following demonstration selection strategy: k-shot prompting, where k is the number of classes. A demonstrative example from each class is randomly selected from the optimization set and represented in the prompt. FuRud significantly improves accuracy and COBias over ICL baselines, as shown in Table \ref{tab:kshot}.

Compared to the 1-shot strategy (Table \ref{tab:comparison}), the k-shot strategy provides a different starting point for FuRud. For example, the average ICL accuracy by k-shot (61.9\%) is slightly larger than that obtained by 1-shot (59.4\%), and average COBias (25.6\%) is smaller than 1-shot (40.5\%). FuRud boosts average accuracy to 73.5\% and reduces COBias to 13.0\%. In conclusion, different example selection strategies provide different starting points to optimize, on which FuRud consistently improve.

\section{More Discussions}
\label{appdix:morediscuss}

\noindent \textbf{We have a different motivation from traditional post-hoc corrections.} Some may argue that ensuring equitable accuracies across all classes is a well-studied problem in standard machine learning classifiers. It is worth emphasizing that the per-class prediction accuracy imbalance should be treated within their particular context. The accuracy bias in ICL outputs stems from completely different causes than the unequal class accuracies observed in potentially overfitted traditional classifiers, where the former is rooted in prompts and the LLMs, and the latter arises from class imbalance of supervised training data. That's why our method is particularly applied to ICL's output token class probabilities, pinpointing specific patterns and applying precise, targeted corrections.

In the future, more versatile rules can be explored, and we may also examine the tradeoff between the accuracy and rule complexity. Simpler rules are easier to understand, but the transformations may fail to catch the intricate interactions between class predictions. More complex rules may have better modeling capabilities, but they are harder to read. In addition, this work focuses on evaluating text classification, and we will extend interpretable ICL debiasing to more language tasks, modalities, and model architectures.


\end{document}